\documentclass[letterpaper, 10 pt, conference]{ieeeconf}  

\IEEEoverridecommandlockouts                              

\usepackage{tcolorbox}
\usepackage{float}
\usepackage{setspace}
\usepackage[dvipsnames,table]{xcolor}
\usepackage{graphicx}
\usepackage{booktabs}
\usepackage{amsmath}
\usepackage{multirow}
\usepackage{makecell}
\usepackage{csquotes}
\usepackage{diagbox}
\usepackage{bm}
\usepackage{xspace}
\usepackage{subcaption}
\usepackage{listings}
\usepackage{wrapfig}

\definecolor{jsonbg}{RGB}{248,248,248}
\definecolor{jsonframe}{RGB}{204,204,204}
\definecolor{jsonkey}{RGB}{0,0,255}
\definecolor{jsonstring}{RGB}{163,21,21}
\definecolor{jsonnumber}{RGB}{0,128,0}
\definecolor{jsonboolean}{RGB}{128,0,128}
\definecolor{jsonnull}{RGB}{128,128,128}
\definecolor{jsonbracket}{RGB}{64,64,64}

\lstdefinelanguage{json}{
    basicstyle=\ttfamily\scriptsize,
    backgroundcolor=\color{jsonbg},
    frame=single,
    framesep=5pt,
    framerule=0.5pt,
    rulecolor=\color{jsonframe},
    xleftmargin=10pt,
    xrightmargin=10pt,
    breaklines=true,
    breakatwhitespace=true,
    showstringspaces=false,
    literate=
        *{:}{{{\color{jsonbracket}{:}}}}{1}
        {,}{{{\color{jsonbracket}{,}}}}{1}
        {\{}{{{\color{jsonbracket}{\{}}}}{1}
        {\}}{{{\color{jsonbracket}{\}}}}}{1}
        {[}{{{\color{jsonbracket}{[}}}}{1}
        {]}{{{\color{jsonbracket}{]}}}}{1}
        {0}{{{\color{jsonnumber}{0}}}}{1}
        {1}{{{\color{jsonnumber}{1}}}}{1}
        {2}{{{\color{jsonnumber}{2}}}}{1}
        {3}{{{\color{jsonnumber}{3}}}}{1}
        {4}{{{\color{jsonnumber}{4}}}}{1}
        {5}{{{\color{jsonnumber}{5}}}}{1}
        {6}{{{\color{jsonnumber}{6}}}}{1}
        {7}{{{\color{jsonnumber}{7}}}}{1}
        {8}{{{\color{jsonnumber}{8}}}}{1}
        {9}{{{\color{jsonnumber}{9}}}}{1},
    morestring=[b]",
    stringstyle=\color{jsonstring},
    morekeywords={true,false,null},
    keywordstyle=\color{jsonboolean}
}
\usepackage{url}
\usepackage{hyperref}
\hypersetup{colorlinks}

\definecolor{background}{RGB}{240,240,240}
\definecolor{keyword}{RGB}{0,0,255}
\definecolor{string}{RGB}{163,21,21}
\usepackage{titletoc}

\newcommand{\D}{\emph{desire}\xspace}
\newcommand{\In}{\emph{intention}\xspace}
\newcommand{\B}{\emph{belief}\xspace}
\newcommand{\V}{\emph{value}\xspace}

\title{\LARGE \bf
Who and Where Am I? Embodied Cognition-Aware Humanoid Agents
}

\author{Fan Lu\textsuperscript{1,2}, Wentao Zhu\textsuperscript{3}, Kwan-Yee Lin\textsuperscript{4}, Wayne Wu\textsuperscript{5}, Yizhou Wang\textsuperscript{6}, Guang Chen\textsuperscript{1,2}
\thanks{\textsuperscript{1}Tongji University~\textsuperscript{2}Shanghai Innovation Institute~\textsuperscript{3}Eastern Institute of Technology~\textsuperscript{4}University of Michigan~\textsuperscript{5}University of California, Los Angeles~\textsuperscript{6}Peking University
}}

\begin{document}

\twocolumn[{%
\renewcommand\twocolumn[1][]{#1}%
\maketitle
\begin{center}
\centering
\vspace{-3ex}
 \includegraphics[width=0.95\linewidth]{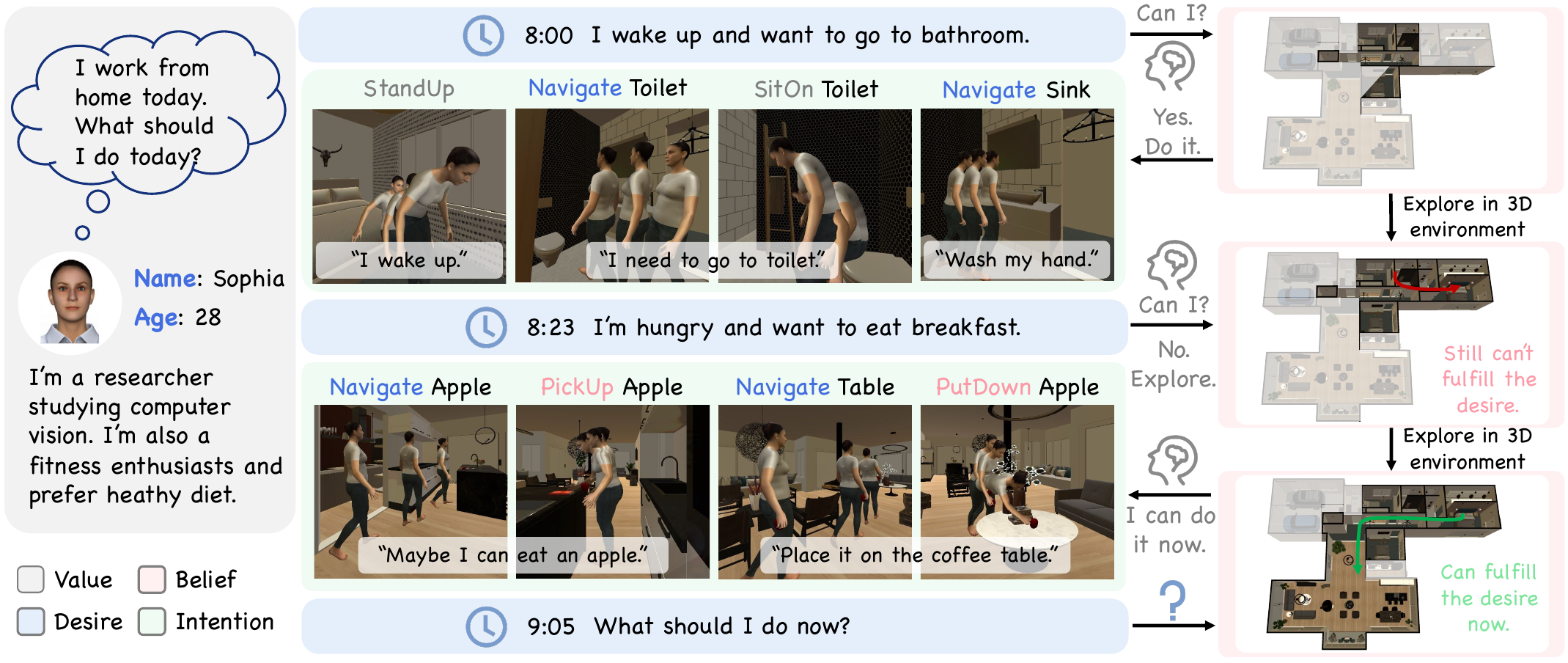} 
\vspace{-1ex}
\captionof{figure}{
\textbf{EmbodiedHuman.} We propose an embodied cognition-aware framework to simulate natural daily behaviors of humanoid agents. With its own personality and preferences (\textit{values}), a humanoid agent, when introduced into an unfamiliar environment, continuously generates temporary \textit{desires} that drive the agent to explore the surroundings while updating its \textit{beliefs} about the environment. Accordingly, the agent formulates specific \textit{intentions} to fulfill its \textit{desires}, which are translated into embodied actions, such as {\textcolor{NavyBlue}{navigation}}, {\color{gray}static object interaction}, {\color{pink}movable object interaction}, and free-form motions.
}\label{fig:teaser}
\vspace{-1ex}
\end{center}
}]

\makeatletter
\begingroup
\renewcommand{\@makefnmark}{}
\footnotetext{
    \textsuperscript{1}Tongji University~
    \textsuperscript{2}Shanghai Innovation Institute~
    \textsuperscript{3}Eastern Institute of Technology, Ningbo~
    \textsuperscript{4}University of Michigan~
    \textsuperscript{5}University of California, Los Angeles~
    \textsuperscript{6}Peking University
}
\endgroup
\makeatother

\thispagestyle{empty}
\pagestyle{empty}

\begin{abstract}

The ability to simulate humanoid agents with human-like daily behavior is a crucial step toward developing intelligent robotic systems, as these systems are expected to operate and interact in human-centric environments. However, current humanoid agents in simulation environments typically emphasize either high-level reasoning or repeated low-level motion execution. As framed by the concept of embodied cognition, humanoid behavior simulation demands capturing the intricate interplay between internal cognitive states, external environments, and physically grounded motion behavior. In this work, we investigate how to endow humanoid agents with such embodied cognitive intelligence.
We propose {\textit{EmbodiedHuman}}, a unified embodied cognitive architecture that tightly couples a structured ``Mind’’ module with generative motor control to drive purposeful behavior in interactive 3D environments. The framework comprises three novel components: $1)$ a cognition-inspired Mind structure, which models and modularize high-level reasoning and decision-making through key causal variables (value, belief, desire, and intention);  $2)$ a generative-based action execution module, which translates internal intentions into embodied movements, enabling physically grounded interactions; and $3)$ an exploration module, which empowers the agent to actively explore the unknown environment and update its mental states via feedback of actions.
Our approach demonstrates a flexible way to populate any simulation scenes with humanoid agents, enabling them to continuously adapt, learn, and evolve their behavior and show dynamic and natural human-like interactions in the long horizon.
Extensive experiments demonstrate the flexibility and scalability of our method as a solid baseline in simulating individualized, daily-level behaviors in unknown environments. Videos are available at~\url{https://embodiedhuman.github.io/}.

\end{abstract}

\section{INTRODUCTION}

Modern robot-training simulators (\textit{e.g.}, Isaac Gym~\cite{makoviychuk2021isaac} and Habitat~\cite{puighabitat}) increasingly support humanoid embodiments and interaction physics, but the behavior of “humanoid agents” is modeled as short-horizon, task-driven, and largely repetitive, which diverges from real human behavior and thereby bias downstream learning and evaluation.
To create genuinely humanoid agents that are helpful for downstream robot training, 
we must simulate the \emph{complexities} of \emph{human behavior}, arising from the interplay between internal mental states and external world. An embodied humanoid agent requires cognitive abilities where the mind drives the body to interact with the environment, and in turn, these interactions shape the mind. This aligns with the philosophical notion of \emph{embodied cognition}, where cognition is not merely a product of the brain but emerges through the agent's physical interactions with environments~\cite{wilson2002six,varela2017embodied}.

Nonetheless, implementing such a framework is far from straightforward. The core challenge lies in simulating the complex, dynamic nature of an agent's internal mental states and their interplay with an external world that is ever-changing, partially observable, and revealed incrementally through embodied interaction. Traditional systems, such as VirtualHome~\cite{puig2018virtualhome} and Watch-And-Help~\cite{puig2020watch}, rely on predefined, rule-based action banks and heuristic search algorithms to retrieve humanoid actions.
In contrast, a truly humanoid agent must exhibit multi-level autonomy: adapting its behaviors and motions dynamically based on evolving mental states shaped by continuous feedback from the environment.
This allows us to flexibly populate simulation scenes with humanoid agents that display natural behaviors, as long as we provide an empty scene and a humanoid profile. These populated scenes can serve as important platforms for various downstream tasks, including human-AI interaction~\cite{puighabitat}, robot learning~\cite{makoviychuk2021isaac}, and embodied AI learning~\cite{kolve2017ai2}.

To achieve this, we introduce an embodied cognitive architecture, {\textit{EmbodiedHuman}}, that seamlessly integrates a high-level cognitive~\textit{``Mind''} concept with low-level motor execution~(Fig.~\ref{fig:teaser}), allowing agents to interact with environments based on their own mental and physical states. 

\begin{figure}
\centering
    \includegraphics[width=0.43\textwidth]{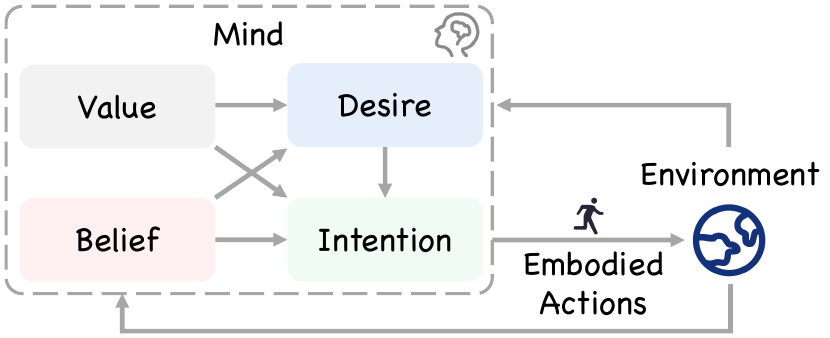}
    \vspace{-1ex}
    \caption{\textbf{Embodied cognitive architecture.} The conceptual workflow of proposed embodied cognitive architecture, shaping the \textit{Mind} and its dynamic interaction with the \emph{Environment} through embodied actions.}
    \label{fig:mind}
    \vspace{-4ex}
\end{figure}

Specifically, grounded in foundational theories in psychology and cognitive science~\cite{bratman1987intention,rao1995bdi,wooldridge2003reasoning}, we structure the~\textit{``Mind''} via modularization through four causal variables: \emph{value}, \emph{belief}, \emph{desire}, and \emph{intention}, it offers a principled framework for self-directed reasoning and action in humanoid agents. Concretely, within the~\textit{Mind}, \emph{value} refers to an individual’s intrinsic motivations, preferences, or long-term goals; \emph{belief} reflects the agent’s perception of the world, which may be accurate or inaccurate; \emph{desire} encapsulates transient or context-dependent aspirations that influence behavior, distinguishing itself from long-term values; and \emph{intention} is the specific action plan chosen to fulfill these \emph{desires}, typically determined through a concrete decision-making process based on the agent's \emph{beliefs} and \emph{desires}. These causally interconnected elements collectively shape the agent’s rich and adaptive behavior, as illustrated in Fig.~\ref{fig:mind}.

While such a structured \textit{Mind} concept grounds the internal cognitive states of the agent, key challenges remain: {\emph{1)} how can the agent physically enact its {\emph{intentions}} through its {\emph{body}}, interact with the {\emph{environment}}, and {\emph{2)} in turn, allow these interactions to shape and update its mental states? To address this, we introduce two complementary interaction modules that operationalize the cognitive architecture, enabling bidirectional adaptation between cognition and embodied experience. The first is the \emph{action execution} module, which translates the agent’s \emph{intentions} into a sequence of executable embodied actions. Leveraging large-scale human motion data~\cite{guo2022generating}, we develop a spatial-aware motion diffusion model to enable controllable motion generation with spatial control, supporting diverse and versatile actions that span multiple levels of granularity, including navigation, object interaction, and free-form human motions. The second is the \emph{exploration} module, which enables the agent to purposefully explore its surroundings and continuously update its mental states (\emph{i.e.}, \emph{beliefs} and \emph{desires}) based on environmental feedback. Specifically, we propose an exploration strategy that directs the agent towards unknown areas based on its current desires, the frontiers~\cite{yamauchi1997frontier} of the scene, and scene understanding.

To summarize, we model humanoid agents via an embodied cognition-aware framework, beyond separate motion synthesis or mind modeling, enabling the population of simulation scenes with humanoid agents exhibiting natural and diverse behaviors. Our technical innovations can be concluded as follows: 
1) We propose EmbodiedHuman, enabling the humanoid agent to continuously adapt, learn, and evolve its behavior based on its own characteristics and interactions with the environment.
2) We design an action execution module that leverages a spatial-aware motion diffusion model to achieve versatile and controllable motion generation, enabling rich environmental interactions. 
3) We develop an exploration module that allows the agent to actively explore the environment, gather information, and establish a feedback loop between its mind and the environment. 
4) Experiments reveal the emergence of diverse, lifelike behaviors, such as picking its favorite foods, exercising, dancing, or simply relaxing in a room, just like any of us in our daily lives. 

\section{Related Works}
\label{sec:related}

\noindent \textbf{Human Behavior Simulation.} Human behavior simulation has long been studied across domains. Early works~\cite{shao2005autonomous,funge1999cognitive} simulate human behavior by defining rules for human-environment interactions and generating motion through basic motor skills. However, these approaches were limited to highly abstracted environments, and the rule-based methodology constrained both behavioral diversity and motion quality.
For behavior planning, cognitive architectures like SOAR~\cite{laird2019soar} and ACT-R~\cite{anderson2014atomic} model human-like decision-making and learning, while the BDI (Belief-Desire-Intention) model~\cite{rao1995bdi} simulates rational agents driven by cognitive states. In virtual environment platforms, VirtualHome~\cite{puig2018virtualhome} simulates 3D household tasks with atomic actions, and Watch-And-Help~\cite{puig2020watch} enables collaborative behaviors, yet remain rule-based with fixed atomic actions, yielding coarse and non-adaptive behaviors that miss fine-grained variations (\textit{e.g.,} grabbing while walking). BEHAVIOR~\cite{srivastava2022behavior} offers a benchmark for realistic human activities, and Habitat 3.0~\cite{puighabitat} provides an advanced simulation platform with accurate humanoid simulation and collaborative tasks. V-IRL~\cite{yang2024virl} drives humanoid agents to interact with urban scenes, while it lacks physical embodiment. 
We advance these efforts with an embodied cognitive architecture that enables agents to autonomously explore, plan, and execute embodied actions in environments, using a feedback loop to update their mind for diverse, adaptive behavior in unfamiliar settings.

\noindent \textbf{Conditional Motion Generation.} 
Generating human motion from various conditions has attracted widespread attention~\cite{guo2022generating,tevet2023human,chen2023executing,zhu2023human,motionpercept2024}. 
Recent advances such as MDM~\cite{tevet2023human, chen2023executing} leverage diffusion models for motion generation. 
Although achieving high-quality motion generation, prior methods typically employ relative motion representation, which undermines spatial controllability, as it lacks a sense of global positions. Several methods attempt to introduce spatial control based on MDM~\cite{karunratanakul2023guided,xie2024omnicontrol}, 
while high control precision remains challenging under the relative representation. 
To address this, we propose a spatial-aware motion diffusion model to incorporate spatial information, thereby enhancing control precision.

\noindent \textbf{Human-Scene Interaction.} 
One line of works trains motion control policies through reinforcement learning (RL) and imitation learning to drive humanoid robot to navigate~\cite{rempeluo2023tracepace,cheng2024navila} and interact with environment~\cite{tessler2024maskedmimic,zhao2023synthesizing,peng2021amp}. 
However, these approaches generally focus on short-term motion and are limited by predefined motion patterns or overfitting to specific environments, overlooking the cognitive nature of humanoid agents.
The other focuses on scene-aware motion generation~\cite{hassan2021stochastic,cen2024generating,jiang2024scaling,jiang2024autonomous}. 
For example, TRUMANS~\cite{jiang2024scaling} expands human-scene interaction dataset and trains an autoregressive motion diffusion model to generate motion based on action label. 
Despite remarkable progress, previous approaches generally assume a {\textit{known}} environment and generate separate motion sequences. Moreover, they synthesizes motions primarily based on external instructions, overlooking the internal mental states of humanoid agents.

\section{Methodology}

\begin{figure*}[htbp]
    \centering
    \includegraphics[width=1.\linewidth]{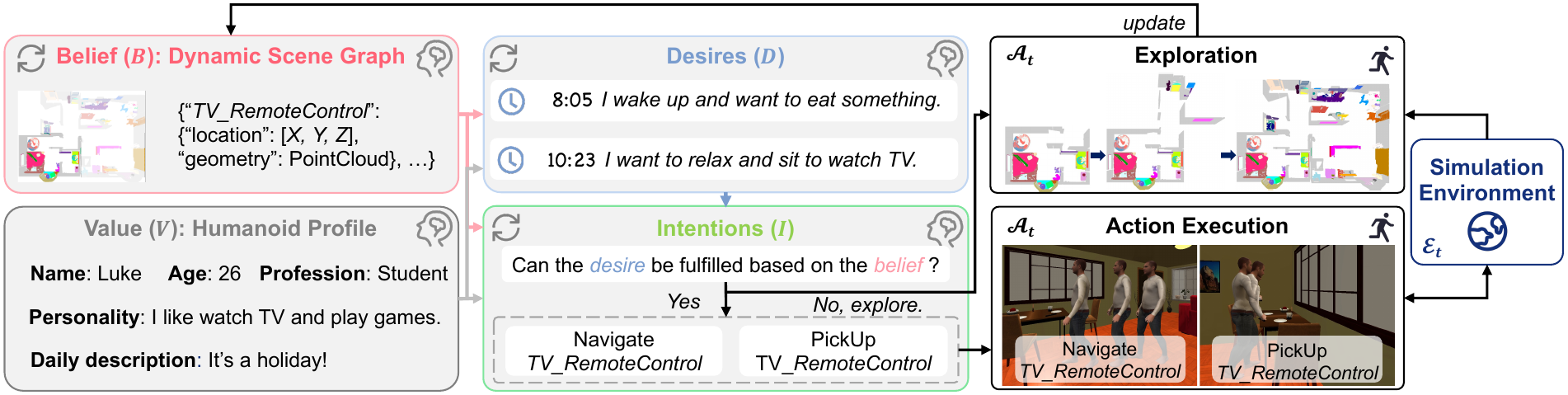}
    \vspace{-4ex}
    \caption{\textbf{Framework of our EmbodiedHuman.} To simulate embodied humanoid agents in unfamiliar environments (Sec.~\ref{sec:formulation}), we design an {embodied cognitive architecture, {\textit{EmbodiedHuman}}, shaping the mind with \textit{value}, \textit{belief}, \textit{desire}, and \textit{intention} (Sec.~\ref{sec:cog_arch}), and coupling cognition with actions for embodied interaction}. Starting from the \textit{value} (humanoid profile) and the \textit{belief} (dynamic scene graph), we leverage an LLM to generate temporary \textit{desires}, which are translated into \emph{intentions}. Two interactive modules are designed to enable embodied \emph{action execution} and dynamic environment \emph{exploration} (Sec. \ref{sec:modules}).
    }
    \label{fig:framework}
    \vspace{-4ex}
\end{figure*}


\subsection{Embodied Cognitive Architecture}\label{sec:formulation}
To populate a simulated scene with humanoid agents exhibiting natural behaviors, we define their behavior using
a triplet $(\mathcal{M}_t, \mathcal{A}_t, \mathcal{E}_t)$, where $t$ is the simulated time. $\mathcal{M}_t$ represents the \emph{Mind} at time $t$, \emph{i.e.}, the internal cognitive states, including \emph{value}, \emph{belief}, \emph{desire}, and \emph{intention}; $\mathcal{A}_t$ represents the \emph{Action} taken by the agent at time $t$, reflecting its embodied interaction with the environment; and $\mathcal{E}_t$ represents the \emph{Environment} at time $t$, \emph{i.e.}, the external world that the agent perceives and interacts with.
Simulated time $t$ can be associated with real-world time (\textit{e.g.,} 8:30 AM) when performing daily activities.

%
 We formulate the interaction between the three components via the following relationships inspired by \emph{embodied cognition}, which emphasizes that cognitive processes emerge from continuous interactions between the mind, body, and environment: $(1)$ The \emph{Action} $\mathcal{A}_t$ of the agent is directly determined by its internal cognitive state, $\mathcal{M}_t$, and the external environment, $\mathcal{E}_t$, via $\mathcal{A}_t = \pi(\mathcal{M}_t, \mathcal{E}_t)$, where $\pi$ is the policy function that maps $\mathcal{M}_t$ and $\mathcal{E}_t$ to a specific action, reflecting the joint influence of both internal mind and external world on the behavior. $(2)$ The~\emph {Environment} $\mathcal{E}$ evolves as a result of the action. Specifically, the new environmental state $\mathcal{E}_{t+1}$ is determined by the previous environment $\mathcal{E}_t$ and the action $\mathcal{A}_t$. Formally, $\mathcal{E}_{t+1} = \mathcal{T}(\mathcal{E}_t, \mathcal{A}_t)$, where $\mathcal{T}$ models how the environment changes in response to actions. $(3)$~The \emph{Mind} $\mathcal{M}_{t+1}$ is updated based on the outcome of the action $\mathcal{A}_t$ interacting with the environment $\mathcal{E}_{t}$, \emph{i.e.}, $\mathcal{M}_{t+1} = \psi(\mathcal{M}_t, \mathcal{A}_t, \mathcal{E}_{t})$, where $\psi$ is a function that models the evolution of the mental states, incorporating feedback from the actions taken and changes in the environment. 

These relationships model the dynamic interaction and feedback loop between the agent's mind, body (actions), and environment, reflecting key insights of \emph{embodied cognition}. Concretely, the environment transition function $\mathcal{T}$ represents the objective evolution of the external environment, which is typically specified by the environment itself. Our primary focus, therefore, is on the implementation of the policy function $\pi$ and the mind update function $\psi$ to more effectively simulate dynamic human behavior through the interplay of the mind, body, and environment. 

\subsection{Mind}\label{sec:cog_arch}

\textit{Mind} consists of four interconnected modules (\V, \B, \In and \D), structured through causal relationships that model human decision-making process. 

\noindent \textbf{Value.} The agent’s \V ($V$) represents its intrinsic motivations, preferences, and goals. These \emph{values} remain relatively stable over the timescale we study, typically measured in hours or days, and guide the agent's overall behavior in the environment. We define the agent’s \V via a character profile, describing attributes such as profession, personality, and other characteristics.

\noindent \textbf{Belief.} The agent’s \B ($B_t$) represents its perception of the environment at time $t$. The \B is updated based on observations the agent makes as it interacts with the environment. In our framework, {\textbf{the agent does not have access to oracle-like global scene information}}; instead, it {\textbf{perceives the environment from an egocentric view}} and dynamically forms its \B through continuous interaction. Thus, \B may be accurate or biased at any given moment, but is gradually refined over time as the agent gathers more information from its surroundings. 

To represent this evolving \B, we employ a dynamically updated 3D scene graph composed of the objects the agent has explored. Each object is characterized by properties such as its 3D location, geometric information, and optional affordances, as illustrated in Fig.~\ref{fig:framework}. 
%
To achieve open-world perception in unknown environments,
we employ Grounded-SAM~\cite{ren2024grounded,kirillov2023segment,liu2023grounding} to extract semantic information from egocentric RGB observations, which is projected into 3D space using depth to generate 3D instance-level point clouds.
As the agent interacts with the environment, the scene graph is progressively updated. This process can be formulated as:
\begin{equation}
\setlength\abovedisplayskip{0.1cm}
\setlength\belowdisplayskip{0.1cm}
B_{t+1} = \mathcal{F}_B(B_t, \mathcal{E}_{t+1}),
\end{equation}
where 
$\mathcal{F}_B$ denotes the belief update function, which is implemented by estimating the overlap between newly perceived instances and existing ones, and then merging them. 
%


\noindent \textbf{Desire.} The agent’s \D ($D_t$) reflects its immediate, context-driven goals at time $t$. Each \emph{desire} can be mapped to a scheduled conceptual time to enable human-readable time representation. While \V represents long-term motivations, \D is more specific and dynamic, shaped by the agent’s current situation and needs. For example, a \D might manifest as ``\textit{ I want to grab some food}'' or ``\textit{I want to take a break}'' at the moment.  The generation of a particular \D is closely influenced by the agent’s \V and evolving \B about the environment, which can be expressed as:
\begin{equation}
\setlength\abovedisplayskip{0.1cm}
\setlength\belowdisplayskip{0.1cm}
D_t = \mathcal{F}_D(V, B_t).
\end{equation}

\noindent \textbf{Intention.} The agent’s {\textit{intention}} ($I_t$) represents a concrete, actionable plan formed to achieve its goals at time $t$. While \D expresses what the agent wants (\emph{e.g.}, ``\textit{I want to eat}''), \textit{intention} addresses the question of how the agent intends to fulfill that \D (\emph{e.g.}, ``\textit{I will go to the kitchen and cook a steak}''). In this way, {\textit{intention}} serves as a bridge between \emph{desire} and action, translating abstract goals into specific steps. Shaped by \V, \D, and the current environmental understanding (\emph{i.e.}, \B), this process can be formulated as:
\begin{equation}
\setlength\abovedisplayskip{0.1cm}
I_t = \mathcal{F}_I(V, D_t, B_t).
\end{equation}

\subsection{Interaction Modules}\label{sec:modules}
We propose two interaction modules to enable the agent to execute physically grounded \emph{Actions} in the environment while establishing a dynamic feedback loop between its internal \emph{Mind} and the external \emph{Environment}. By tightly coupling these \emph{Actions} with perception and cognition, the framework ensures that the agent automatically generates coherent and natural motions in the environment.
Specifically, if the agent determines, based on its current \B, that its \D can be satisfied, the \emph{action execution module} translates \emph{intentions} into embodied actions. 
Otherwise, the \emph{exploration module} actively drives the agent to explore the environment, guided by the agent’s \B and \D, and, in turn, updates them as new information is gathered from the surroundings.

\subsubsection{Action Execution}\label{sec:action}
The action execution module translates \emph{intentions} into executable humanoid actions, requiring precise spatial control and free-form motion generation for accurate scene interaction and diverse behaviors. 

While employing RL to control humanoid actions is a straightforward solution, it often yields limited and unnatural motion patterns and can easily overfit to specific scenes. Leveraging motion generation models with priors from large-scale human motion data offers a more promising path, either via standalone~\cite{tevet2023human} or scene-aware~\cite{jiang2024scaling} manners. 

However, such pre-trained models often fail to satisfy both properties—either lacking effective environmental interaction or struggling to generalize across diverse environments due to data limitations. Based on the above observations, we develop a spatial-aware motion diffusion model (SA-MDM), which
enhances spatial control capability through two design levels: {\textit{spatial-aware motion representation}} and {\textit{gradient-based motion control}}. Such designs improve the agent's ability to interact with the environment while utilizing large-scale scene-agnostic human motion data. As shown in Fig.~\ref{fig:action_execution}, SA-MDM enables precise motion control via both texts and spatial cues, allowing the agent to perform specific, context-aware actions. We unfold the details below. 

\begin{figure}
    \centering
    \includegraphics[width=0.48\textwidth]{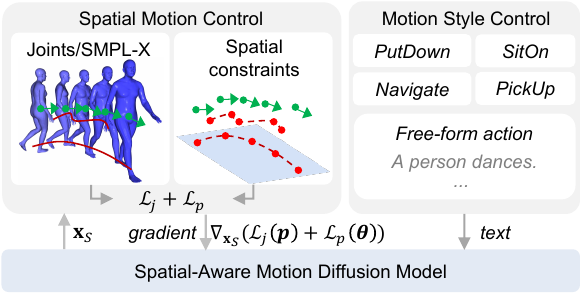}
    \vspace{-4ex}
    \caption{\textbf{Overview of the proposed action execution module.} Text controls the motion style of Spatial-Aware Motion Diffusion Model (SA-MDM). The diffusion process is further guided by gradients from spatial constraints, where red curves indicate sparse joint locations (taking root and hand joints as examples) and green arrays are global orientations.}
    \label{fig:action_execution}
    \vspace{-4ex}
\end{figure}
\noindent\textbf{SA-MDM.} Most previous human motion generation models follows the relative, local motion representation proposed in HumanML3D, meaning that the joint representation of the current frame is relative to the previous frame. This representation diminishes the ability for precise global position control~\cite{karunratanakul2023guided}, which is crucial for ensuring physically plausible interactions with the environment. To address this, we primarily use global joint positions $\bm{p}$ as the motion representation. However, this representation is limited to skeleton-level control and does not account for the agent's physical volume. Consequently, relying solely on joint control is inadequate for preventing penetration with surrounding objects. Thus, we propose a hybrid spatial-aware motion representation that fuses joint positions with SMPL-X parameters $\bm{\theta}$. We follow MLD to perform diffusion in a latent space (compressed with VAE) for better efficiency and motion quality. The two representations are tightly coupled throughout the generation process to ensure coherence, where joints govern the global position and posture, while the mesh vertices derived from SMPL-X parameters provide the necessary body shape information for penetration avoidance.

\noindent\textbf{Motion Control.} With SA-MDM, we can control the motion in a spatial-aware manner through two key components. First, we regulate the global joint positions to perform specific environment-aligned actions by designing tailored joint loss functions $\mathcal{L}_j$ for different types of controls (\emph{e.g.}, joint position, heading angle, \emph{etc}). Second, we reduce potential human–scene penetration by computing a penetration loss $\mathcal{L}_p$ between mesh vertices and the dynamically updated scene occupancy. Thanks to our spatial-aware motion representation, both forms of control can be seamlessly applied by incorporating the guidance directly into the denoised latent code, which is formally expressed as:
\begin{equation}
\setlength\abovedisplayskip{0.1cm}
\setlength\belowdisplayskip{0.1cm}
    \tilde{\mathbf{x}}_s = \mathbf{x}_s - \alpha \nabla_{\mathbf{x}_s}(\mathcal{L}_j(\bm{p})+\mathcal{L}_p(\bm{\theta})), (\bm{p},
\bm{\theta}) = \mathcal{D}(\mathbf{x}_s) \end{equation}
where $\mathcal{D}(\cdot)$ represents the decoder of the VAE in SA-MDM, $\tilde{\mathbf{x}}_s$ and $\mathbf{x}_s$ denote the perturbed and original latent estimates at time $s$, and $\alpha$ represents the gradient scale. 
Using this strategy, we can effectively control agents' interaction with the external environment in a unified manner. 
Additionally, we enable text-based motion control, allowing for freeform style adjustments. These texts leverage the cognitive context, making the motion more flexible and personalized. 

\noindent\textbf{Training with Multi-Source Data.} To enhance motion quality for complex interactions, we augment the HumanML3D~\cite{guo2022generating} dataset with additional human-object interaction data from SAMP~\cite{hassan2021stochastic} and GRAB datasets~\cite{taheri2020grab}.

\begin{wrapfigure}{r}{0.70\linewidth}
  \vspace{-1.2em}
  \centering
  \includegraphics[width=\linewidth]{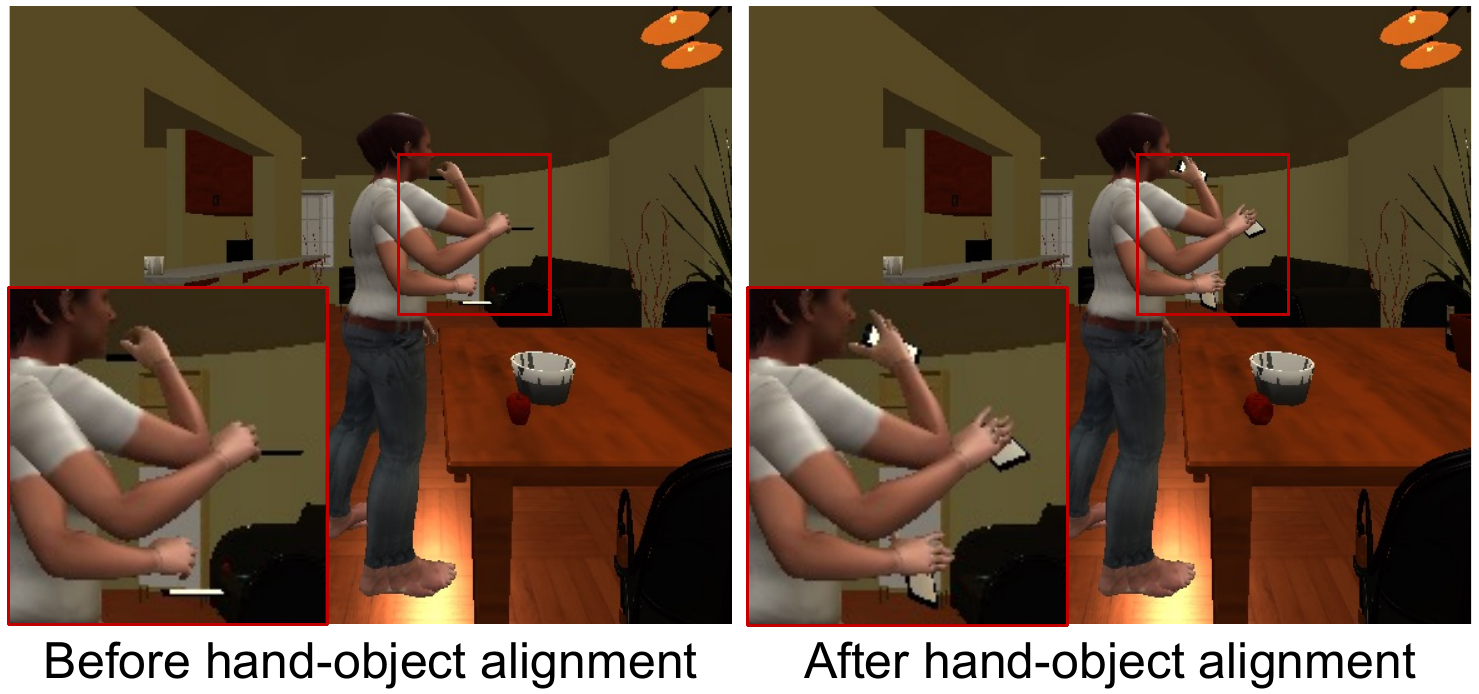}
  \vspace{-2ex}
  \caption{Effectiveness of hand-object alignment strategy.}
  \label{fig:hand_object}
  \vspace{-1.2em}
\end{wrapfigure}
\noindent\textbf{Hand-object Alignment.} 
Realistic hand-object interaction is an essential component for populating simulations with high-quality humanoid agents. To achieve this,
we leverage GRIP~\cite{taheri2024grip} to generate hand poses based on hand and object geometry, as shown in Fig.~\ref{fig:hand_object}. Since GRIP does not provide object's poses, we determine it via a two-step process: first, we perform an initial alignment using the hand's surface normals, and second, we track the object's pose frame-by-frame with hand's rigid transformation by registrating the hand vertices. 


\subsubsection{Exploration}\label{sec:explore}
Notably, since the agent dynamically gathers information and forms its \B about the environment through continuous interactions (as discussed in Sec.~\ref{sec:cog_arch}), it may not always be able to form a specific {\textit{intention}} that fulfills its \D (\emph{e.g.}, when a required object is in another room, and the agent is unaware of it). In such cases, we propose an exploration module with a desire-driven strategy, where exploration itself becomes a form of {\textit{intention}}. This enables the agent to actively seek new information in response to its unmet \D, and further update its mental states.

Specifically, the agent needs to distinguish between known and unknown areas, and identify regions for further exploration. In practice, we compute two distinct maps during exploration: explored map and obstacle map. By analyzing their differences, we detect frontiers (\emph{i.e.}, boundaries between explored and unexplored spaces)~\cite{yamauchi1997frontier}. However, these frontiers lack semantic information. To bridge this gap and connect the frontiers with human-like cognition, we employ an LLM to assess the exploration priority of each frontier based on the current \D and descriptions of surrounding object instances. For an intuitive example, if the current \D is to go to the bathroom, a frontier near a sink would more likely be useful and therefore prioritized for exploration. Nonetheless, there are times when the \D simply cannot be fulfilled within the environment. We terminate the process when no frontiers remain or when the exploration for a single \D exceeds a certain number of attempts. At that point, we update the agent's \D based on the feedback, just as a real human would. This process reflects how active interaction with the environment influences the agent's mind, including its \B and \D. 

\section{Experiments}


\subsection{Experimental Setup}\label{sec:exp_setup}

We model $V$, $B_t$, $D_t$, and $I_t$ using natural language and implement $\mathcal{F}_D$ and $\mathcal{F}_I$ by prompting a pretrained LLM (GPT-4o~\cite{hurst2024gpt} in our experiments), considering the complexity of human mental states. 
We leverage an open source virtual platform, Habitat 3.0~\cite{puighabitat}, and use SMPL-X~\cite{pavlakos2019expressive} as the skeleton representation.
We select 10 scenes from HSSD~\cite{khanna2024habitat} and AI2-THOR dataset~\cite{kolve2017ai2}. We define 10 humanoid profiles encompassing different professions, personalities, daily goals, \emph{etc}. We then randomly assign humanoid agents to scenes, creating 20 human-scene combinations. 


\subsection{Daily Humanoid Behavior Simulation}
\label{sec:results}

\begin{figure}
    \centering
    \includegraphics[width=0.93\linewidth]{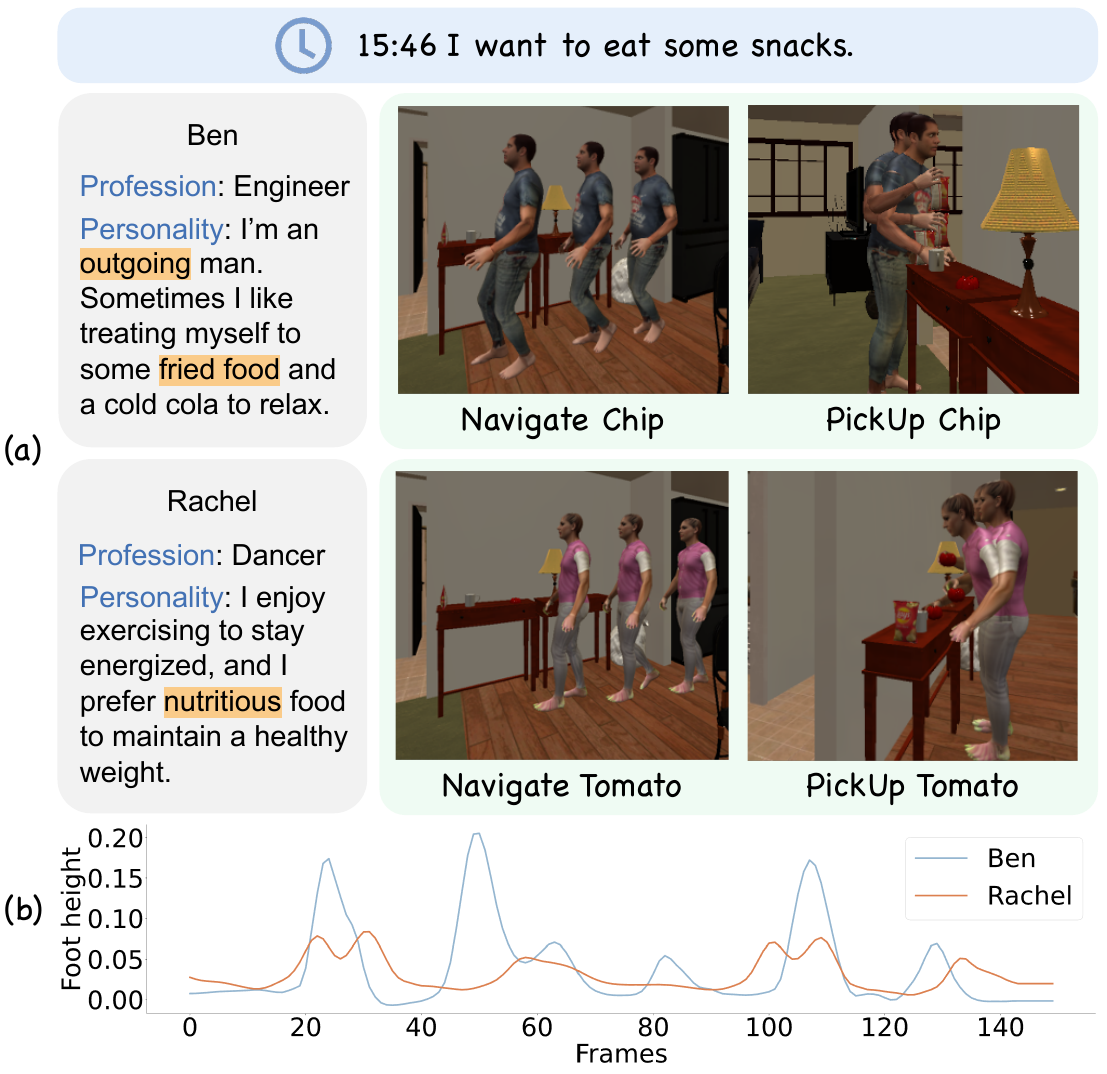}
    \vspace{-1ex}
    \caption{\textbf{\textit{Value} influences agent's behavior.} (a) Two agents (\emph{i.e.}, Ben and Rachel) tend to eat different foods (Chip or Tomato) according to their \textit{values}. (b) As an outgoing individual, Ben’s {\textit{gait}} appears brisker, which is also reflected in the foot height visualization at the bottom.}
    \label{fig:value_analysis}
    \vspace{-3ex}
\end{figure}

We investigate two key research questions (RQs) 
to examine if EmbodiedHuman can reflect variations in internal \V and external environment,  while generating diverse, contextually appropriate behaviors that support populating simulation scenes with natural humanoid agents:

\noindent\textbf{RQ1: How does \emph{value} influence agents' behavior?} Agents' behavior is shaped by their \emph{values}, leading to distinct choices even in the same environment. To examine this, we conduct a controlled experiment with two humanoid agents sharing the same \emph{desires} in the same environment. As shown in Fig.~\ref{fig:value_analysis}(a), Ben and Rachel select different foods according to dietary preferences: Ben chooses chips due to his preference for fried food, while Rachel opts for a tomato, valuing nutritional benefits. Additionally, we compare the quantitative gait difference via visualizing the foot height of both agents (Fig.~\ref{fig:value_analysis}(b)). We observe that Ben’s gait is brisker, aligning with his \V of being ``outgoing." 

\begin{figure}
    \centering
        \includegraphics[width=0.93\linewidth]{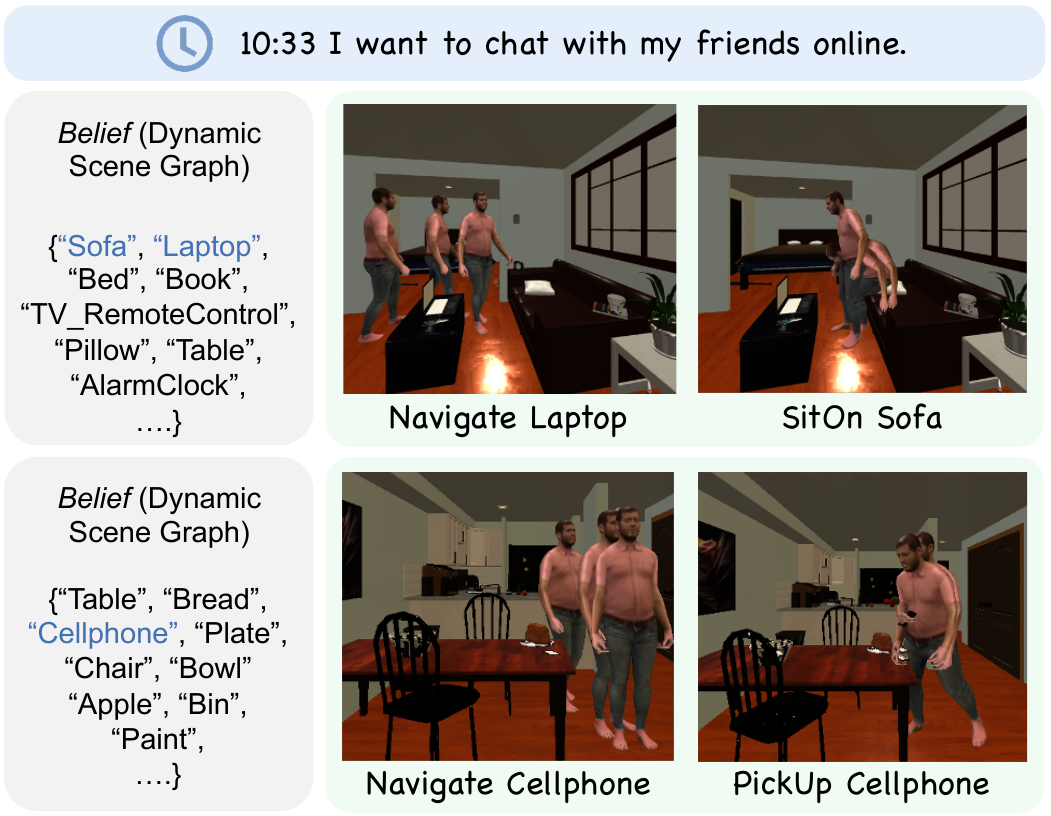}
        \vspace{-1ex}
        \caption{\textbf{\textit{Environment} influences humanoid agents' behavior.} Armin chooses different assets (Laptop or Cellphone) to fulfill his \emph{desire} based on his \textit{belief} about the environment.}
        \label{fig:belief_analysis}
        \vspace{-4ex}
\end{figure}

\noindent\textbf{RQ2: How does the \emph{environment} influence agents' behavior?} The external environment plays a crucial role in shaping the cognition and behavior. As shown in Fig.~\ref{fig:belief_analysis}, we place the agent Armin into two different environments with the same \emph{desire} and observe different decisions. To fulfill the \emph{desire} to ``chat with friends online,'' Armin chooses either a cellphone or a laptop, depending on availability in the environment. By perceiving environmental cues, the agent forms different \emph{beliefs}, leading to diverse yet reasonable behaviors. We further examine how scene dynamics shape agents' behavior, as illustrated in Figs.~\ref{fig:dyna_eat} and~\ref{fig:dyna_cola}.

\begin{wrapfigure}{r}{0.65\linewidth}
  \centering

  \includegraphics[width=\linewidth]{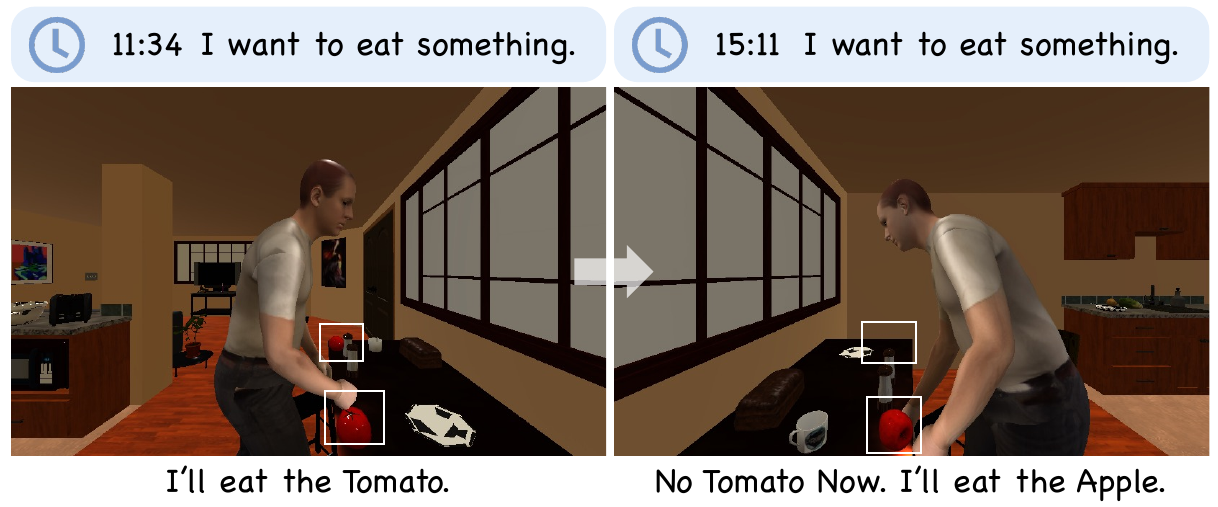}
  \vspace{-3ex}
  \caption{The agent chooses to eat Apple when Tomato has been eaten.}
  \label{fig:dyna_eat}

  \vspace{0.8em}
  \includegraphics[width=\linewidth]{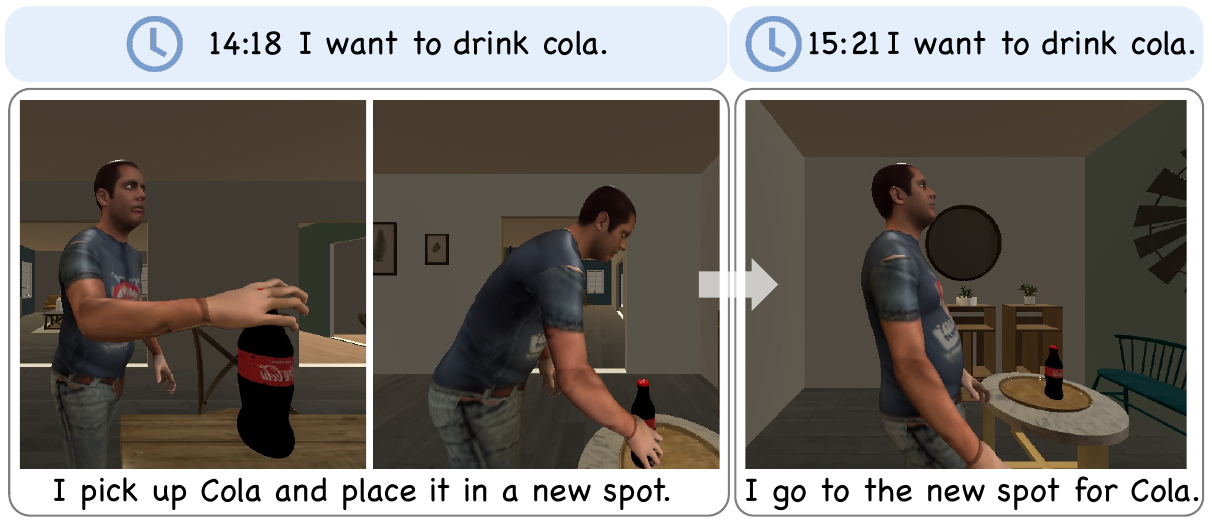}
  \vspace{-3ex}
  \caption{The agent navigates to a different place due to Cola's location change.}
  \label{fig:dyna_cola}

  \vspace{-1.2em}
\end{wrapfigure}
\noindent\textbf{{Simulate a Day of Humanoid Agents.}} Based on the ability of EmbodiedHuman, we can populate simulation environments with dynamic and diverse humanoid behaviors at a daily scale.
Figs.~\ref{fig:visual_fiona} and \ref{fig:visual_diego} depict a simulated day in the lives of two agents, Fiona and Diego. They have distinct professions and personalities, which lead them to make different choices throughout the day. For example, Fiona enjoys chatting on the phone and dancing, while Diego prioritizes working out to maintain energy. Results show that our pipeline can generate various behaviors based on \emph{values}, allowing for rich and meaningful environmental interactions. This is achieved through mind-environment interaction, along with diverse humanoid actions, including navigation, static/dynamic interactions (\emph{e.g.}, \textit{SitOn} and \textit{PickUp}), and open-vocabulary motions (\emph{e.g.}, \textit{Dance Cha-Cha}).

\begin{wrapfigure}{r}{0.60\linewidth} 
  \vspace{-1.2em}
  \centering
  \includegraphics[width=\linewidth]{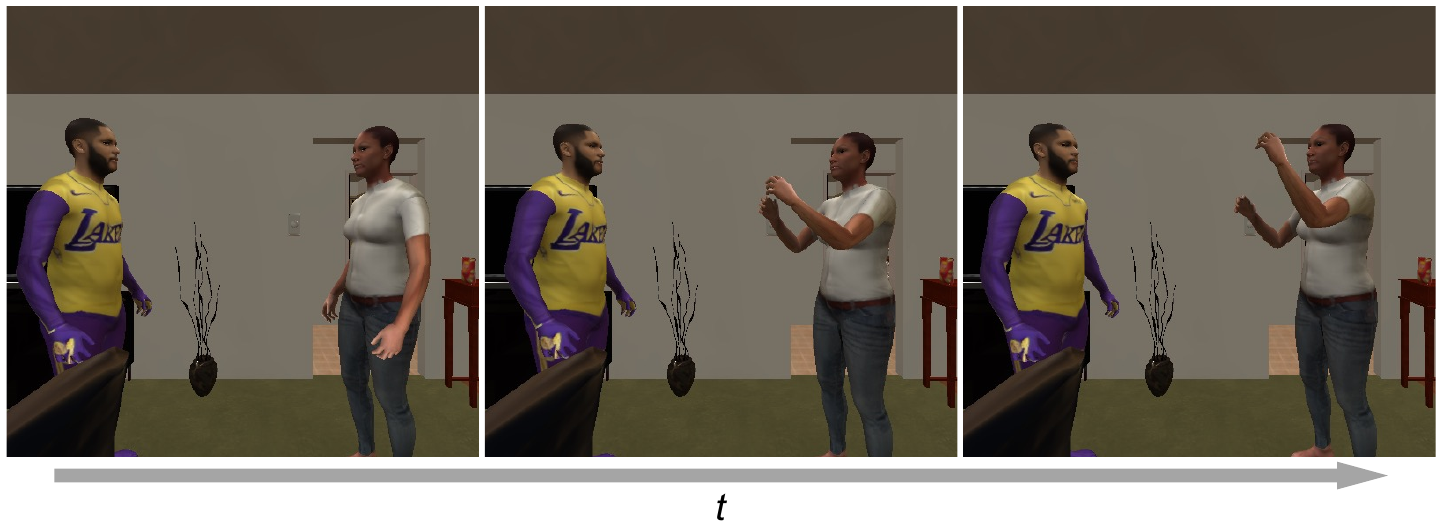}
  \vspace{-3ex}
  \caption{Multi-agent interaction.}
  \label{fig:multihuman}
  \vspace{-1.2em}
\end{wrapfigure}
\noindent\textbf{Extension to Multi-agent Interaction.} Our framework is primarily designed for single-agent behavior modeling; nevertheless, it can be readily extended to multi-agent scenarios by incorporating \textit{Interaction} as an additional \textit{intention}, as shown in Fig.~\ref{fig:multihuman}.

\begin{figure*}[htbp]
    \centering
    \includegraphics[width=0.96\linewidth]{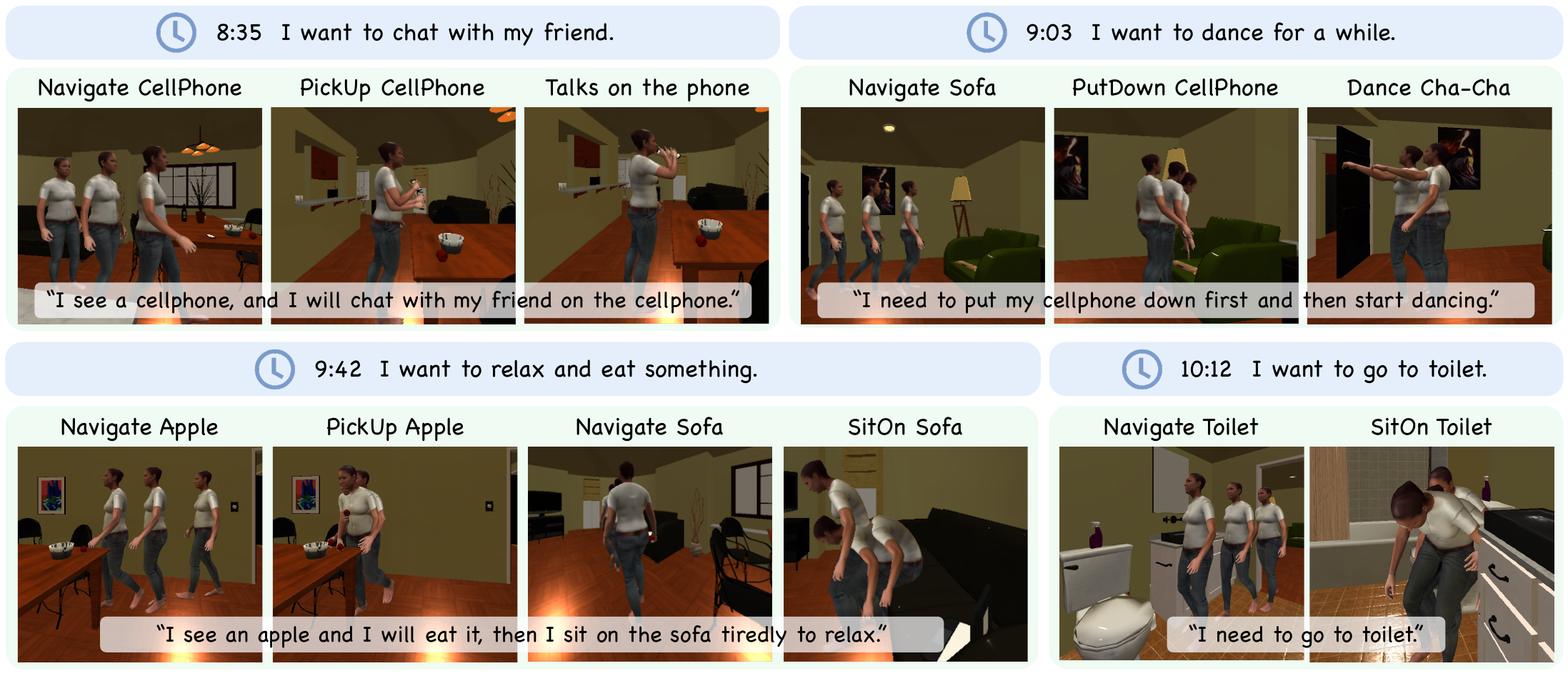}
    \vspace{-2ex}
    \caption{\textbf{A day of Fiona.} Fiona is a manager working at a trading company. She is a social butterfly and is passionate about dancing. Today is a holiday. We generate long-term daily behaviors of Fiona: She starts the day by chatting with her friend on the phone and then dances to wake up herself. Thereafter, she feels tired and sits down to eat an apple. Then, she needs to go to toilet. The simulated behaviors reflect the \emph{value} of Fiona and the \emph{belief} from the environment.}
    \label{fig:visual_fiona}
    \vspace{-2ex}
\end{figure*}

\begin{figure}[htbp]
    \centering
    \includegraphics[width=1.\linewidth]{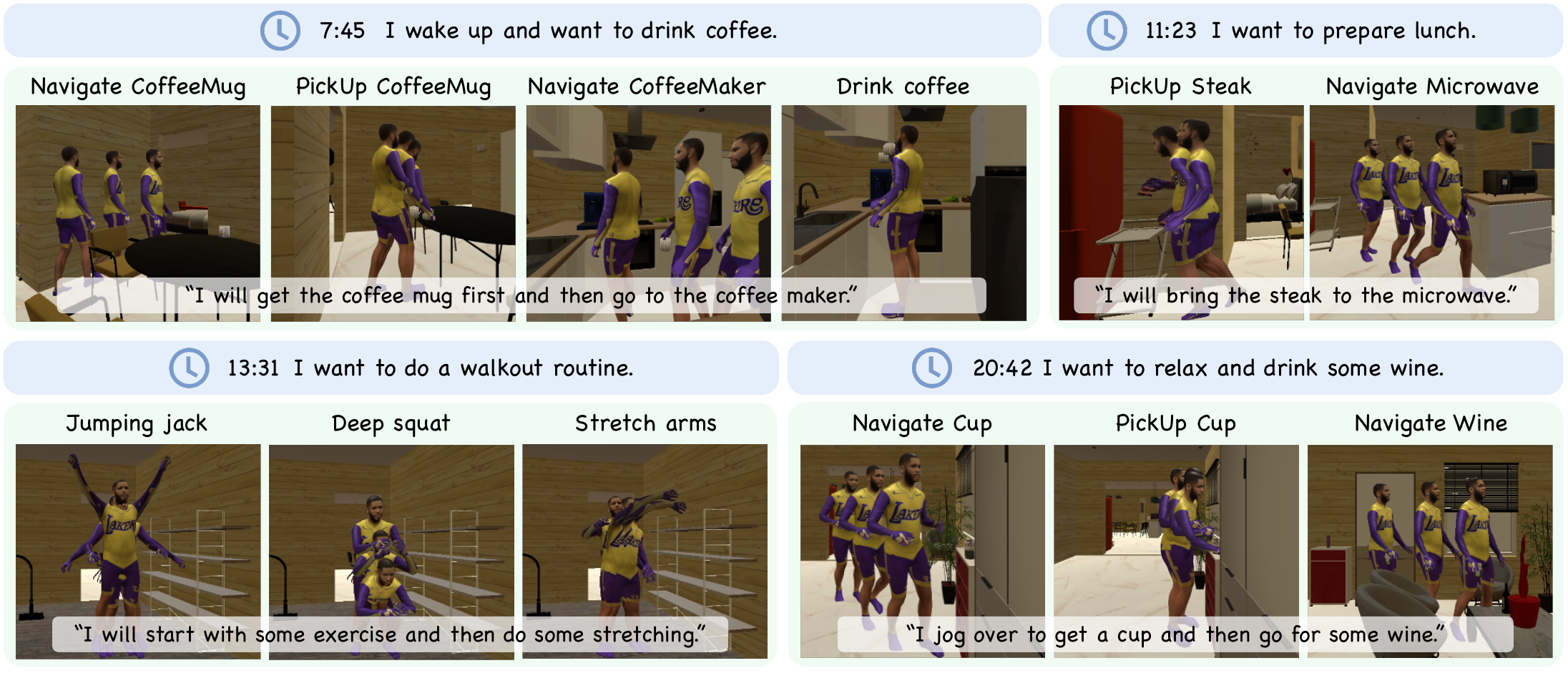}
    \vspace{-3ex}
    \caption{\textbf{A day of Diego.} As an athlete, he enjoys drinking coffee to stay alert and occasionally has alcohol to relax. Today, he needs to exercise. He starts the day with coffee to energize himself and chooses steak for lunch to meet his high-protein preference. Then, he completes a workout routine to meet his daily goal, and unwinds with a glass of wine in the evening. Better Zoom in for details.}
    \label{fig:visual_diego}
    \vspace{-3ex}
\end{figure}

\subsection{Evaluation of Exploration Module}\label{sec:exploration_exp}



\begin{wraptable}{r}{0.65\linewidth} 
  \vspace{-1.2em}
  \footnotesize
  \centering
  \setlength\tabcolsep{3pt}
  \caption{Exploration module evaluation.}
  \label{tab:exploration}
  \vspace{-1.0em}
  \begin{tabular}{lcc}
    \toprule
    Method & Daily Plan. $\uparrow$ & Desire Plan. $\uparrow$ \\
    \midrule
    w/o Exploration & 0.05 & 0.54 \\
    w/o Desire & 0.45 & 0.91 \\
    Ours & \textbf{0.65} & \textbf{0.95} \\
    \bottomrule
  \end{tabular}
  \vspace{-1.0em}
\end{wraptable}
We introduce two metrics to assess the planner's ability to generate valid \emph{intentions} that fulfill \emph{desires}: Desire Plan Success Rate (\emph{Desire Plan.}) and Daily Plan Success Rate (\emph{Daily Plan.}). The former measures the ratio of individual \emph{desires} that can be fulfilled based on the agent’s \emph{belief}, while the latter evaluates the success rate of fulfilling all \emph{desires} within a day.
We experiment with removing \D guidance and ablating the exploration module entirely. Table \ref{tab:exploration} shows that both \emph{Daily Plan.} and \emph{Desire Plan.} drop significantly without exploration. In this scenario, the agent’s ability to achieve its \emph{desires} depends solely on its initial position; if the current \emph{belief} cannot satisfy the \emph{desire}, the plan fails. Additionally, incorporating \emph{desire} further improves success rates by enabling semantically-aware exploration.

\subsection{Evaluation of Action Execution Module}
\label{sec:action_exp}

\begin{table}
    \centering
    \setlength\tabcolsep{0.6pt}
    \footnotesize
    \caption{Evaluation of action execution module. $^*$We adapt the baseline methods to our framework by carefully modifying them to replace our action execution module.}
    \label{tab:daily_action}
    \vspace{-2ex}
    \begin{tabular}{llllll}
    \toprule
        Methods & \makecell{Daily\\Exec.$\uparrow$} & \makecell{Desire\\Exec.$\uparrow$} & \makecell{Action\\Exec.$\uparrow$} & \makecell{Goal\\Dist.$\downarrow$}  & \makecell{Pene.$\downarrow$} \\
    \midrule
        OmniControl$^*$~\cite{xie2024omnicontrol} & 0.05 & 0.64 & 0.83 & 0.313 & 0.039 \\
        TextSceneMotion$^*$~\cite{cen2024generating} & 0.00 & 0.57 & 0.80 & 0.265 & 0.209 \\
        TRUMANS$^*$~\cite{jiang2024scaling} & 0.20 & 0.70 & 0.88 & 0.267 & 0.033 \\
        \midrule
        Ours & \textbf{0.30} & \textbf{0.79} & \textbf{0.91} & \textbf{0.237} & \textbf{0.027} \\
    \bottomrule
    \end{tabular}
    \vspace{-3ex}
\end{table}
We assess the precision of the action execution module using the following metrics: Daily Executed Rate (\emph{Daily Exec.}), Desire Executed Rate (\emph{Desire Exec.}), Action Executed Rate (\emph{Action Exec.}), Goal Distance (\emph{Goal Dist.}), and Penetration Ratio (\emph{Pene.}). Specifically, \emph{Goal Dist.} refers to the distance between the joint and the goal; 
\emph{Action Exec.} measures the rate at which the executed action successfully realizes the intention (\emph{i.e.}, when \emph{Goal Dist.} is below a predefined threshold $\tau$). \emph{Daily Exec.} and \emph{Desire Exec.} represent the proportion of fully successful intentions at the daily and desire levels. For example, \emph{Daily Exec.} requires all intentions within a day to succeed; if any fail, the entire day is considered unsuccessful. 

\noindent\textbf{Comparison with SOTA.} We compare our method against two scene-aware human motion generation methods (\emph{i.e.}, TextSceneMotion~\cite{cen2024generating} and TRUMANS~\cite{jiang2024scaling}) and a spatially controllable motion generation method (\emph{i.e.}, OmniControl~\cite{xie2024omnicontrol}). Specifically, we replace our action execution module with each baseline method, adapting them to fit our scenarios.
As shown in Table \ref{tab:daily_action}, TextSceneMotion exhibits a notably higher penetration ratio, likely due to its object-centric approach, but its dataset limits transfer to new Habitat 3.0 environments. TRUMANS benefits from a larger human-scene interaction dataset, but its success rates remain significantly lower than ours, especially at the desire and daily levels, where errors accumulate across actions. OmniControl employs joint position control for human-scene interaction, but its lower precision leads to much lower success rates. We attribute this to its reliance on relative motion representation.

\section{Conclusion}
We propose EmbodiedHuman, an embodied cognitive architecture to simulate intelligent humanoid agents behavior by integrating ``Mind'' with two interactive modules. We shape the ``Mind'' with four interconnected causal variables: \emph{value}, \emph{belief}, \emph{desire} and \emph{intention}. To achieve physically grounded interactions, we propose an action execution module to generate diverse motions and design a desire-driven exploration module to actively explore unknown environments. Experiments show that EmbodiedHuman supports individualized, daily-level behavior simulation.

\bibliographystyle{IEEEtran}
\bibliography{IEEEabrv}

\end{document}